\documentclass[letterpaper, 10 pt, conference]{ieeeconf}
\usepackage{cite}
\usepackage[left=48pt, right=60pt, top=60pt, bottom=50pt]{geometry}
\usepackage{bm}
\usepackage{amsmath}
\usepackage{amssymb}
\usepackage{graphicx}
\usepackage{xcolor}
\usepackage{float}
\usepackage{dcolumn}
\usepackage{siunitx}
\usepackage{tabularx}
\usepackage{tabulary}
\usepackage{array}
\usepackage{booktabs}
\usepackage{xfrac}
\graphicspath{{./media/}}

\IEEEoverridecommandlockouts
\newcommand{\change}[1]{{\color{black}#1}}

\title{\large\bf A Biomimetic Gaze Control to Restore Head-Neck Movements}

\author{Colin Rubow$^{1,2}$ and Haohan Zhang$^{1,*}$
\thanks{This work was partially supported by the National Institutes of Health (R21EB035378 and R01HD119073), the ALS Association (24-SGP-692), and the National Institute of Occupational Safety and Health (5T42OH008414).}
\thanks{$^{1}$Department of Mechanical Engineering and Robotics Center, University of Utah, Salt Lake City, UT 84112, USA.}
\thanks{$^{2}$Rocky Mountain Center for Occupational and Environmental Health, University of Utah, Salt Lake City, UT 84112, USA.}
\thanks{$^{*}$This is the corresponding author (e-mail: haohan.zhang@utah.edu).}}

\begin{document}

\maketitle

\begin{abstract}
Powered neck exoskeletons are developed to restore head-neck motions for those with neck muscle weakness. Current methods using hand-held input devices to control these robotic systems are unintuitive or non-viable for these populations. Alternatively, the eyes coordinate with the head to accomplish many daily tasks. Therefore, movements of the eyes can be a useful signal to control the neck exoskeletons without needing to use the hands. However, previous control models lack biobehavioral basis as well as means to personalize the control behavior, causing unnatural head-neck movements. This paper presents a novel eye-tracking controller design leveraging the vestibulo-ocular reflex to emulate natural eye-head coordination. 
Personalization tools were also designed, consisting of sliders to allow users to adjust the parameters of this controller. Here, we present the technical design and validation of this system in human users. Compared to the previous eye-tracking model, the new controller provides natural head motions personalized to the user through more natural velocity profiles and a \change{24} percent average increase of accurate motion intent detection. This system provides a powerful platform for studying eye-tracking controllers in the context of personalization and preference, and ultimately restoring the lost head-neck mobility in patient populations.
\end{abstract}


\section{INTRODUCTION}
Powered neck exoskeletons are developed to restore lost head-neck mobility for patients with dropped head syndrome~\cite{zhang2022amyotrophic,demaree2024preliminary}. Patients with this condition cannot maintain an upright head posture for an extended time due to fatigue, often resulting in a ``chin-on-chest'' posture~\cite{brodell2020dropped}. 
Everyday activities such as eating and conversing become extremely taxing for these patients~\cite{kiernan2011amyotrophic,walling1999amyotrophic}. The clinic standard method of using static neck collars to treat this condition is unsatisfactory~\cite{burke2025cervical}; these collars only keep the head in an upright position and thus prohibit any voluntary head movements~\cite{pancani2016assessment, baxter2016evaluating, glazener2014pilot}. As a result, these collars do not help with daily tasks for these patients, which is a critical unmet need.

In contrast, neck exoskeletons use external actuators to enable head-neck movements. Although the mechanical hardware of current neck exoskeletons provides sufficient support to the head~\cite{demaree2024preliminary}, controlling them effectively by the user remains challenging. For example, hand-held devices such as joysticks and keyboards were previously used~\cite{zhang2017active, demaree2023structurally,bales2024six}, but such options are not viable for patients with concurrent poor hand functions, which is common in patients whose dropped head posture results from a neurodegenerative disease like amyotrophic lateral sclerosis~\cite{walling1999amyotrophic,oskarsson2016upper}. Even when the hands are functional for using these hand-held input devices, mapping the hand motion to the head rotations is not an intuitive task; the consequences of making a wrong move could result in significant discomfort.

Surface electromyography has also been explored to assist neck muscles of healthy individuals~\cite{wu20166}; however, further study is necessary as benefits were only shown in a simulation. Additionally, neck muscles are used during activities like speaking and drinking~\cite{matsuo2008anatomy}, which furthers challenges using these muscle signals to control neck exoskeletons. It is also unclear whether such control schemes would be effective for people with neck muscle weakness who may have reduced electromyographic signals.

As the head often coordinates with the eyes to gather visual information~\cite{nakashima2014we}, eye movements can potentially be a useful signal to control neck exoskeletons in a natural and intuitive manner. Two attempts to leverage eye-tracking for control have been implemented. The first utilized a gaze clicking method to control a two degrees of freedom (DoF) head support device attached to a wheelchair through a stationary eye-tracking device~\cite{macdonald2010study}. With this control method, the device can move the head in horizontal or vertical rotations but not at the same time. Because the eye-trackers were fixed to the chair, accuracy decreases as the head moves away from the stationary tracker. This strategy frees the hands, but is unnatural because the eyes must fixate on the screen during the control of the head motions. 

Another attempt used a head-mounted eye tracker to control a wearable neck exoskeleton by moving the head when the gaze is in a predefined region~\cite{chang2020bio}. The controller has four gaze regions (each occupies 90$^\circ$ range partitioned by 45$^\circ$ diagonal lines) acting like virtual buttons to command constant horizontal and vertical head rotation increments, and \change{an elliptical} central region which stops the head rotations. These rotations are performed in real-time while the gaze is outside the central region but, like the previous example, performs only constant-speed horizontal or vertical rotations separately. Diagonal trajectories are therefore performed in zigzag fashions because the gaze constantly moves between the borders of two adjacent regions.

Although utilizing concepts in head-eye coordination like the vestibulo-ocular reflex (VOR), it is unlikely that this control strategy obeys the natural kinematics during coordination behaviors, due to its single directional rotations and fixed speeds. The fixed parameters (speed, shape and size of the central region) also disallow this controller to adapt to potential interpersonal variability. In the rest of this paper, we will refer to this controller as the ``region-based controller''.

To address these limitations, we propose a novel eye-tracking control strategy for powered neck exoskeletons which is aimed to emulate essential head-eye coordination behavior and provides means for personalization. \change{We will refer to this controller as the ``vector controller''.}
Here, our contributions are: (i) design and validation of a simple control model, based on observed head-eye coordination data, to predict head rotations given gaze as the input; (ii) evaluation of this control model implemented on a powered neck exoskeleton with human users; and (iii) justification and preliminary evaluation of a method for personalizing the control model.
These contributions provide the technical foundation for future user experiments to determine optimal, personalized, and effective control strategy for powered neck exoskeletons to restore lost head-neck mobility.

The rest of the paper is organized as follows: section~\ref{sec:design} introduces the insights into developing the \change{vector} gaze control model, and section~\ref{sec:robot} describes the implementation of this control model with a powered neck exoskeleton. We then evaluated the exoskeleton control through a user study which is described in detail in section~\ref{sec:exp}. This is then followed by sections~\ref{sec:result} and~\ref{sec:discuss} which present and discuss the results from the human study. Finally, section~\ref{sec:conclusion} summarizes the main contributions of the paper.

\section{Design of a Novel Gaze Control Model}\label{sec:design}
\subsection{Eye-Head Coordination}
In the literature, gaze is defined as the head angle plus the eye angle relative to the head, or, the direction of sight~\cite{freedman2008coordination}. When a gaze shift is performed, the eyes quickly shift towards the target (saccade). The VOR then enables the counter rotations of the eyes and head so that the gaze remains fixed on the target while the eyes become centered in the head~\cite{tweed1995eye,land2006eye}.

Additionally, the peak velocity of the head during a gaze shift was found to be linearly proportional to the distance that the head would travel to recenter the eyes within the field of view~\cite{freedman2008coordination,barnes1979vestibulo, freedman1997eye}. We refer to this linear proportionality as the ``velocity proportion'' \change{($v_p$)} in this paper. A smaller gaze deviation would therefore require a slower head movement, while a larger gaze deviation would need a faster head rotation during VOR. The specific trajectory of the head varies with each gaze shift; however, it has been shown that the head generally moves towards a gaze target in a straight path, even for targets that are not strictly vertical or horizontal~\cite{tweed1995eye, tweed1997three}.

To enable visual exploration without head movements, there also must exist a region near the center of field of vision that induces no head movement should the gaze be inside it. In this paper, we refer to this region as the ``deadzone''. Little is known, however, about the shape and size of this deadzone region. \change{It was hypothesized in the region-based controller's design that the deadzone should be elliptical in shape~\cite{chang2020bio}; however, justification for this was based on literature on detectable light increment thresholds~\cite{strasburger2011peripheral} rather than the border of head moving induced gaze positions. As a result, the deadzone region in the previous model was a wide elliptical shape with fixed size parameters (24.72$^\circ$ horizontal diameter, and 12.96$^\circ$ vertical diameter).} We \change{also} hypothesized that this region can be modeled using a simple elliptical model with the origin of the ellipse at the center of the field of vision. The rationale was based on that the shape of field of vision is conic~\cite{kheirandish2012arabic} and that there exists variation of head contribution for gazing at vertical and horizontal visual stimuli~\cite{freedman2008coordination, glenn1992violations}. \change{Unlike previous work~\cite{chang2020bio}, our elliptic model in this work will be verified with human participant data, and will be personalized to the user.} An elliptic deadzone model is also simple because it requires only two independent parameters, which is a potential advantage for controller personalization, especially  for manual parameter tuning~\cite{ingraham2022role}.

\subsection{Eye-Tracking Control Algorithm}
\begin{figure}[htb]
    \centering
    \includegraphics*[width=\linewidth]{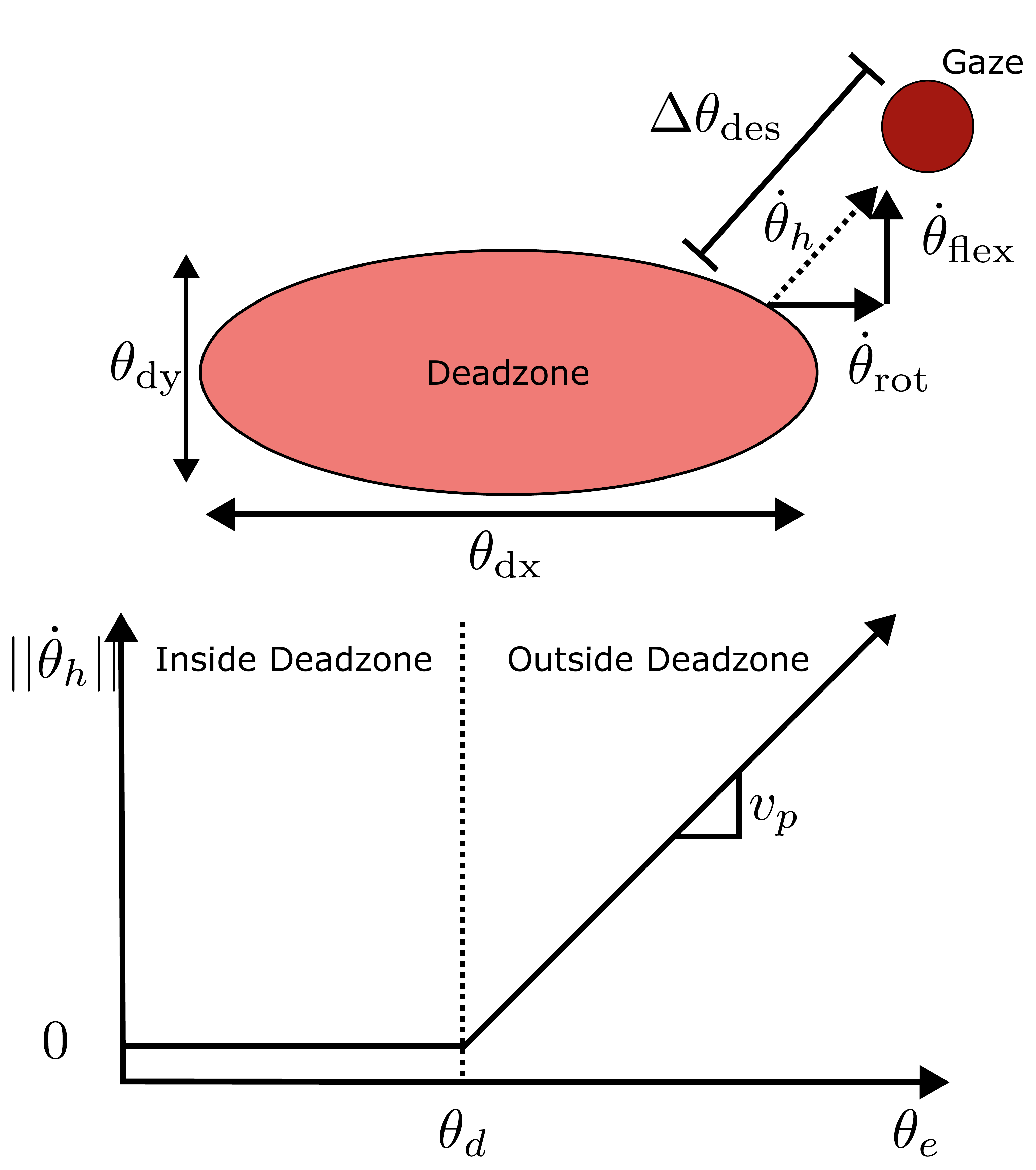}
    \caption{Deadzone and gaze diagram with plot of rotational speed related to gaze shift distance. $\Delta \theta_\text{des}$ is the desired angle of the rotation of the head. \change{$\dot{\theta}_h$} is the velocity of the head which is broken into its pitch and yaw rotational components. $d_x$ and $d_y$ are the lengths of the deadzone elliptical axis where the deadzone is centered in the world camera view. While the gaze is inside the deadzone, the head velocity is 0, but once outside the deadzone, the velocity increases linearly at the rate of $v_p$, the velocity proportion, in the direction of the gaze. \label{fig:control_diagram}}
\end{figure}

We propose a gaze controller model for head movements to incorporate the ``velocity proportion'' and the ``elliptical deadzone'' design principles\change{, the vector controller}. In this model, as illustrated in Fig.~\ref{fig:control_diagram}, when the gaze is within the deadzone, parameterized by the horizontal \change{($\theta_{dx}$)} and vertical diameters \change{($\theta_{dy}$)}, the desired angular velocity of the head \change{($\dot{\theta}_h$)} is zero. When the gaze moves outside the deadzone, the head angular velocity is in the direction pointing from the center of the field of view to the target, while its amplitude \change{($\|\dot{\theta}_h\|$)} is proportional to the shortest distance of the target to the border of the elliptical deadzone \change{($\Delta \theta_{\text{des}}$)} with a constant scaling factor of $v_p$, the velocity proportion. The head angular velocity is then decomposed into two components, \change{$\dot{\theta}_{rot}$ and $\dot{\theta}_{flex}$, which correspond to the head axial rotation about a world fixed vertical axis, followed by the head flexion/extension about the head fixed horizontal axis. Given the two dimensional data from the gaze coordinate, the inference for the third rotation, i.e., the lateral bending of the head, is assumed zero. Therefore, this decomposition strategy is equivalent to the body yaw-pitch-roll sequence.}

Once the gaze is fixated on the target outside of the deadzone and the head moves towards the target, the VOR will be invoked which counter-rotates the eyes back towards the center of the field of vision~\cite{freedman2008coordination}. Using this model, the head will stop its movement when the gaze measured in the head frame hits the border of the deadzone.

For rotation about each axis, (the vertical for \change{axial rotation} and horizontal for \change{flexion/extension}), the control law for a gaze shift can be expressed independently as,
\change{\begin{equation}
\dot{\theta}_h = v_p(\theta_e - \theta_d)\,,
\label{eq:one}\end{equation}}
where $\theta_h$ and $\theta_e$ are the head angle in the inertial frame and the eye angle measured in the head frame, respectively, and \change{$\theta_d$} is a constant which is the deadzone length \change{for that gaze direction}.

The gaze angle in the inertial frame (denoted using $\theta_g$) is the sum of the head angle ($\theta_h$) and eye angle in the head ($\theta_e$). Therefore,
\change{\begin{equation}
\theta_e = \theta_g - \theta_h\,.
\end{equation}}
By \change{substituting this relationship into Equation~(\ref{eq:one})}, we then obtain,
\change{\begin{equation}
\dot{\theta}_h = v_p(\theta_g - \theta_h - \theta_d)\,.
\label{eq:three}\end{equation}}

We now make the substitution,
\change{\begin{equation}
\bar{\theta}_h = \theta_g - \theta_h - \theta_d\,.
\label{eq:four}
\end{equation}}
\change{Assuming $\theta_g$ throughout the VOR process is held constant in a gaze fixation, the time derivative of Equation~(\ref{eq:four}) results in the following:}
\change{\begin{equation}
\dot{\bar{\theta}}_h = -\dot{\theta}_h\,.
\end{equation}}

The control law \change{from Equation~(\ref{eq:three})} can then be written as,
\change{\begin{equation}
\dot{\bar{\theta}}_h = -v_p\bar{\theta}_h\,,
\end{equation}}
which is a first-order ordinary differential equation with the solution,
\change{\begin{equation}
\bar{\theta}_h = \bar{\theta}_{h}(0)e^{-v_pt},
\end{equation}}
where $\bar{\theta}_{h}(0)$ is the initial condition. Unpacking $\bar{\theta}_h$, we obtain the angular trajectory of the head as,
\change{\begin{equation}
\theta_h = \theta_g - \theta_d - (\theta_{e}(0) - \theta_d)e^{-v_pt}\,.
\end{equation}}

\change{Therefore, this control law results in a first order system for gaze shifts with time constant $1/v_p$.} A positive value of $v_p$ also guarantees the system to be bounded input bounded output stable\change{, guaranteeing the head velocity converges to zero as the gaze is centered in the field of view,} which is important to ensure safety when controlling neck exoskeletons. 

\subsection{Controller Comparison in Simulation}

Compared to the previous region-based gaze control model, this new \change{vector} model has multiple advantages. The \change{personalizable} deadzone with an elliptical shape allows the range of local eye exploration (without moving the head) to vary in the horizontal and vertical directions~\cite{freedman2008coordination, glenn1992violations} while having a smooth transition for gaze shifts that are in between\change{, tailored to the user}. Additionally, while the region-based model commands constant velocities in only vertical or horizontal direction at a time, the \change{vector} model is able to rotate the head in any direction at a velocity that is related to the distance of the target, which matches the biological findings~\cite{freedman2008coordination,barnes1979vestibulo}. Unlike the region-based model, this \change{vector} model allows direct head rotations to targets that are not strictly in the vertical or horizontal directions.  

To further illustrate and quantify \change{the greater} ability of the \change{vector} model to bring the head directly to the targets as compared to the region-based model, we performed a computer simulation. 
The angular trajectories of the head resulting from the \change{vector} and region-based controllers were compared in a series of 100 computer simulated gaze shifts. Specifically, 
we measured the ``path efficiency'', defined as the path lengths, of the two models. Shorter path lengths correspond to more efficient head trajectories because they bring the head to the target in a direct fashion like the human do~\cite{tweed1995eye, tweed1997three}. 

In this simulation, the gaze input was assumed to always be fixated on each target. 
To make a fair \change{and simplified trajectory} comparison, the deadzones of both models were set to be an identical circle with a radius of 10$^\circ$. Note that the circular-shaped deadzone is a special case in the \change{vector} model when the vertical and horizontal axes of the ellipse are of the same length. Each gaze shift was performed from the zero \change{flexion} and \change{rotation} position to a gaze target.

\change{Under these conditions, comparing upward gaze shifts is identical to comparing downward gaze shifts and comparing rightward gaze shifts is identical to comparing leftward gaze shifts. Furthermore, since the gaze command borders of the region-based controller are determined by lines of 45$^\circ$ slope~\cite{chang2020bio}, gaze shifts that are symmetrical about these lines will also have identical comparisons. In other words, we only need to compare gaze shifts in an eighth of a circle from directly rightward to 45$^\circ$ upwards.}

\begin{figure}[htb]
    \centering
    \includegraphics*[width=\linewidth]{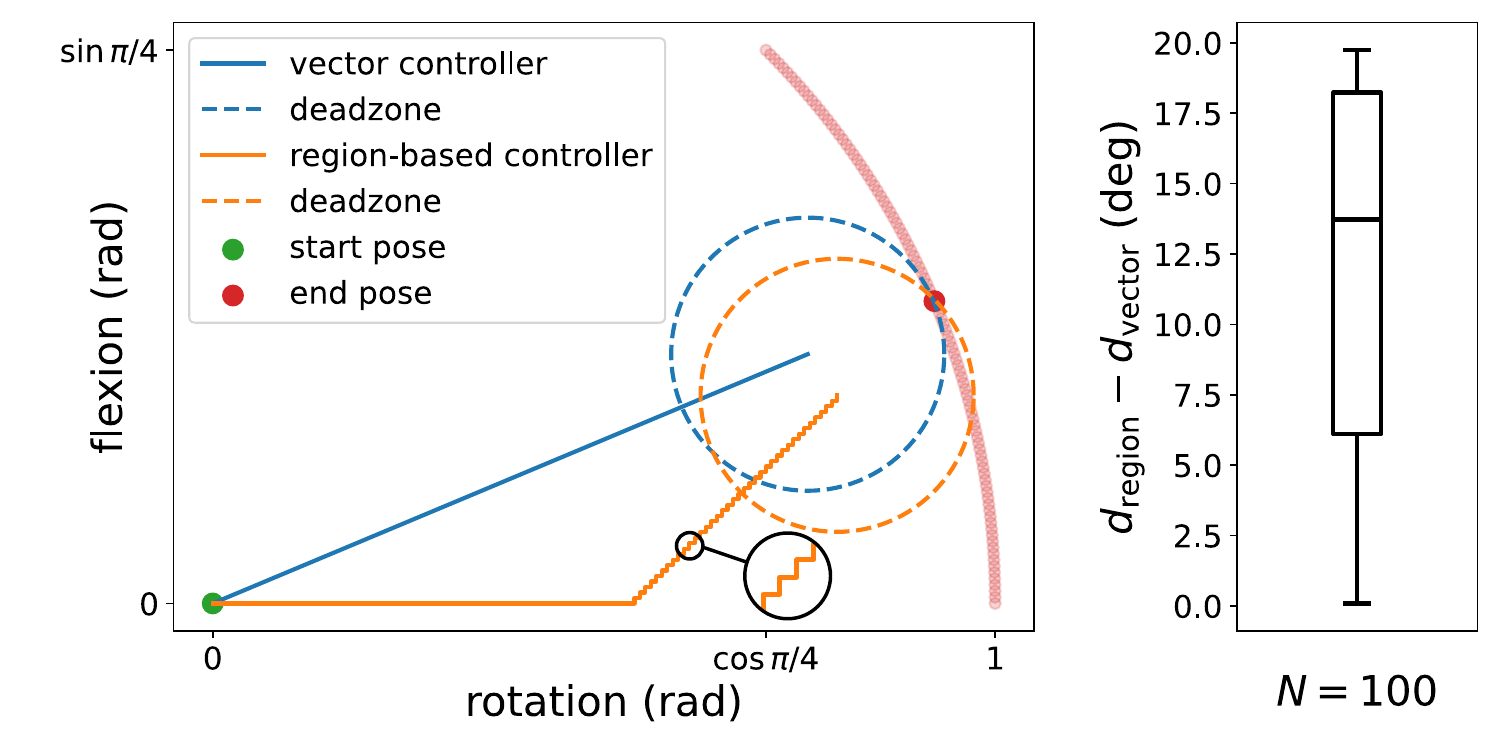}
    \caption{Trajectory comparison results between the \change{vector} and region-based controllers in 100 computer simulated gaze shifts. (left) A typical gaze shift performance. The \change{vector} controller moves straight towards the target, while the region-based controller moves first to the right, then zigzags towards the target (enlarged in the zoomed view). Both controllers stop when the gaze target hits their respective deadzones, the dashed lines of the same color. \change{The 100 gaze targets are distributed from directly rightward to upwards with a slope of 45$^\circ$}. (right) Box plot of the difference of path lengths between both controllers. Higher values correspond to a higher path length performed by the region-based controller as compared to the \change{vector} controller. The median value is $13.7^\circ$. \label{fig:simulation}}
\end{figure}

An example gaze shift is shown in Fig.~\ref{fig:simulation} (left). The angular trajectory of the head resulting from the \change{vector} controller moves directly towards the gaze target, as expected. The region-based controller moves indirectly towards the gaze target with first a horizontal shift due to the gaze point being located in the ``right'' virtual button region~\cite{chang2020bio}. This was followed by a trajectory consisting of short horizontal and vertical zigzag rotations once the gaze target hit the border between the ``up'' and ``right'' virtual buttons in the head frame. 
The differences of trajectory lengths between the two models for all 100 targets are summarized in Fig.~\ref{fig:simulation} (right). The minimum difference is $0^\circ$ when the gaze target is directly horizontal and the maximum difference is $19.7^\circ$ when the gaze target is at an incline of $45^\circ$.

\section{Robotic Integration}\label{sec:robot}
We then implemented \change{the vector} gaze control model with a powered neck exoskeleton to evaluate the model performance in the context of a human-robot system. In this section, we describe the hardware and software implementation of this controller, as well as tools for tuning controller parameters for potential personalization.

\begin{figure}[htb]
    \centering
    \includegraphics*[width=\linewidth]{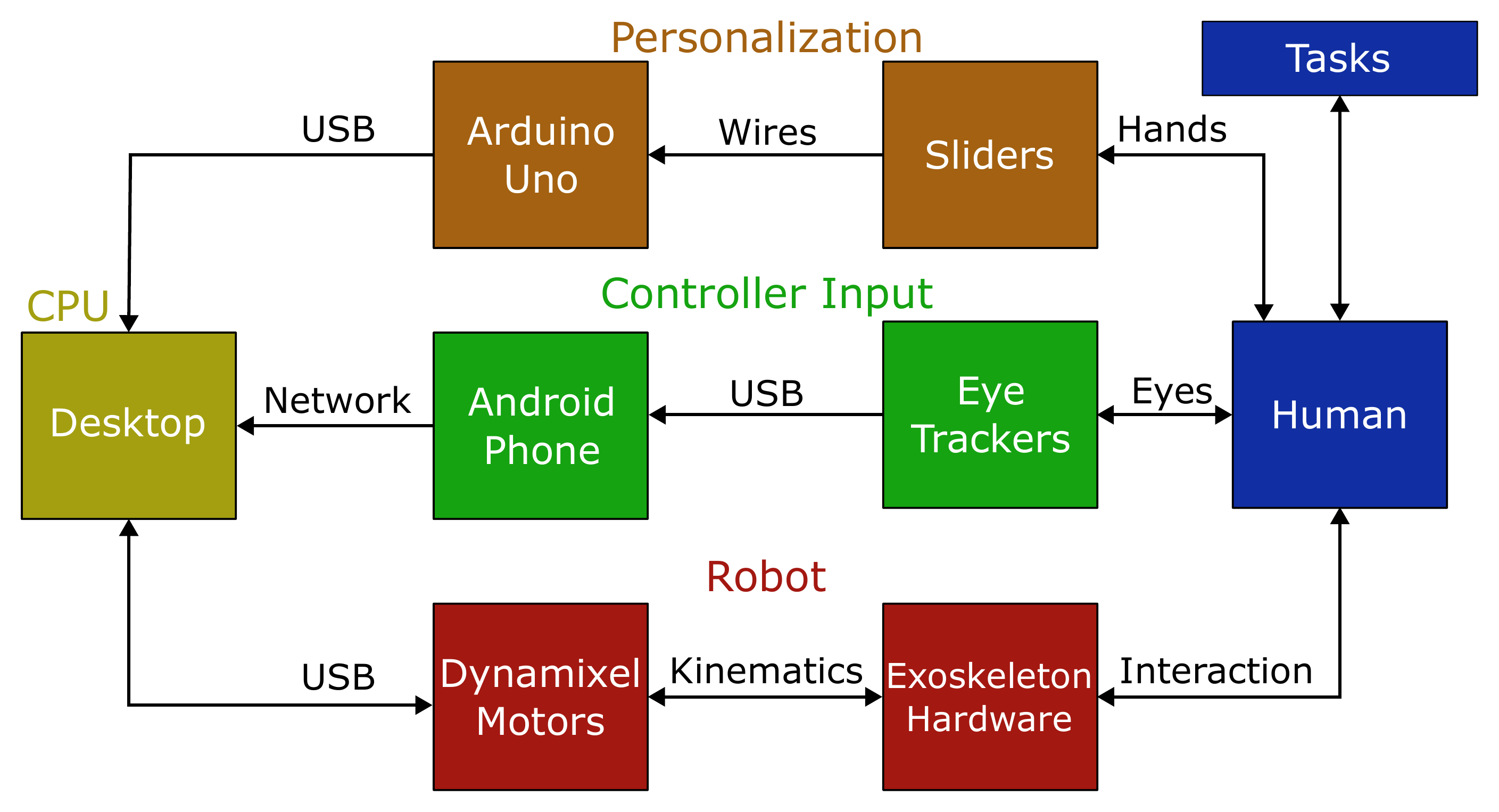}
    \caption{The robotic system. The arrows denote the directional interaction subsystems have with each other. The android phone runs a program that allows gaze data streaming. \label{fig:system}}
\end{figure}

The robotic system contains four main components  (Fig.~\ref{fig:system}): an exoskeleton hardware, a head-mounted eye-tracker, mechanical sliders for manual parameter tuning, and computer programs associated with control software. 

The exoskeleton structure uses a 3-DoF parallel mechanism to allow for the spatial rotations of the head. The geometry of the linkages and joints were optimized to achieve large ranges of rotations of the head relative to the torso~\cite{zhang2017kinematic}. The exoskeleton attaches to the body at the shoulders via pads and straps and on the forehead through an adjustable padded headband. This exoskeleton provides up to 70\% range of motion of the head-neck~\cite{zhang2017kinematic}. Mechanical hard stops were used to prevent the robot from moving past the natural movement range. 

\begin{figure}[htb]
    \centering
    \includegraphics*[width=\linewidth]{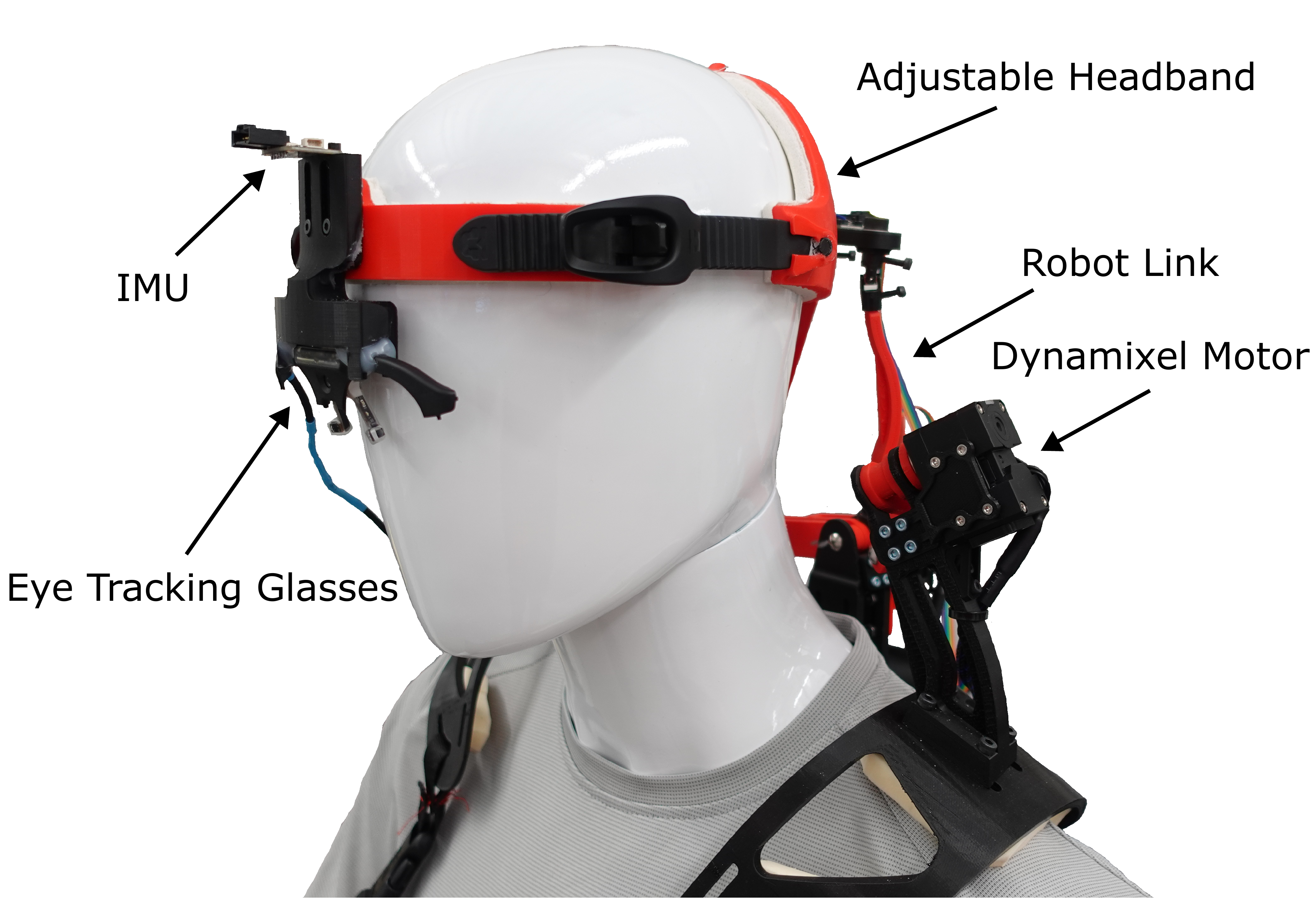}
    \caption{The powered neck exoskeleton with eye-tracking integration. An inertial measurement unit (IMU) was included for the study as an independent measurement of head motions, but it was not used in the robot control. The headband can be easily detached from the rest of the robot by removing a set of fasteners.\label{fig:neck_brace}}
\end{figure}

Improvements made to the original design in~\cite{zhang2019novel} include a more comfortable and form fitting headband along with components for attaching an eye-tracker onto the headband (Fig.~\ref{fig:neck_brace}). The attachments can be adjusted to fit the user and provide clear camera coverage of the eyes. Three servo motors (Dynamixel XM430-w350-r, ROBOTIS Inc. South Korea) were used for actuation, which directly communicate with the computer over a USB connection through an RS-485 serial communication protocol. For simplicity, we did not utilize the portability of the exoskeleton, but instead tethered it to a desktop computer and wall outlet power for data acquisition and processing in the ensuing user study.

A commercial head-mounted eye-tracker (Neon, Pupil Labs, Berlin, Germany) was used to measure eye angles. This tracker provides accurate and robust out-of-the-box tracking for this variable. Images of the eyes and environment are fed as input to a black box machine-learned model trained on a diverse and expansive dataset. This tracker can achieve an accuracy of 1.8$^\circ$, does not require calibration, is resistant to slippage, blinking, and different lighting environments, and can be worn over glasses~\cite{baumann2023neon}. Gaze data was sampled at a rate up to 200~Hz and filtered with an exponential moving average of smoothing factor 0.1.

During robot control, once the desired head velocity is determined through \change{the vector} control model, the motor velocities are calculated using the velocity kinematics of the exoskeleton~\cite{zhang2019novel}, 
\begin{equation}\change{\dot{\bm{\theta}_h}}=\bm{J}\bm{\omega} \,, \end{equation}
where \change{$\dot{\bm{\theta}_h}$}, $\bm{J}$, and $\bm{\omega}$ are velocities of the servo motors, the velocity Jacobian matrix of the robot, and the desired head angular velocity, respectively.

The servo motors were programmed to track velocities, which provides a smoother and quieter control experience than the positional control used in~\cite{chang2020bio}. \change{If the brace pushes too hard on the head, current limits in the motors cause the motors to shut off for safety.} We \change{also} limited the orientation of the head to an effective range of 40$^\circ$ flexion, 10$^\circ$ extension, and 30$^\circ$ \change{rotation} in both left and right axial rotations. When one of these limits is reached, the head angular velocity component in that direction is set to 0. The robot may then slide along the border in the allowed directions.

\begin{figure}[htb]
    \centering
    \includegraphics*[width=\linewidth]{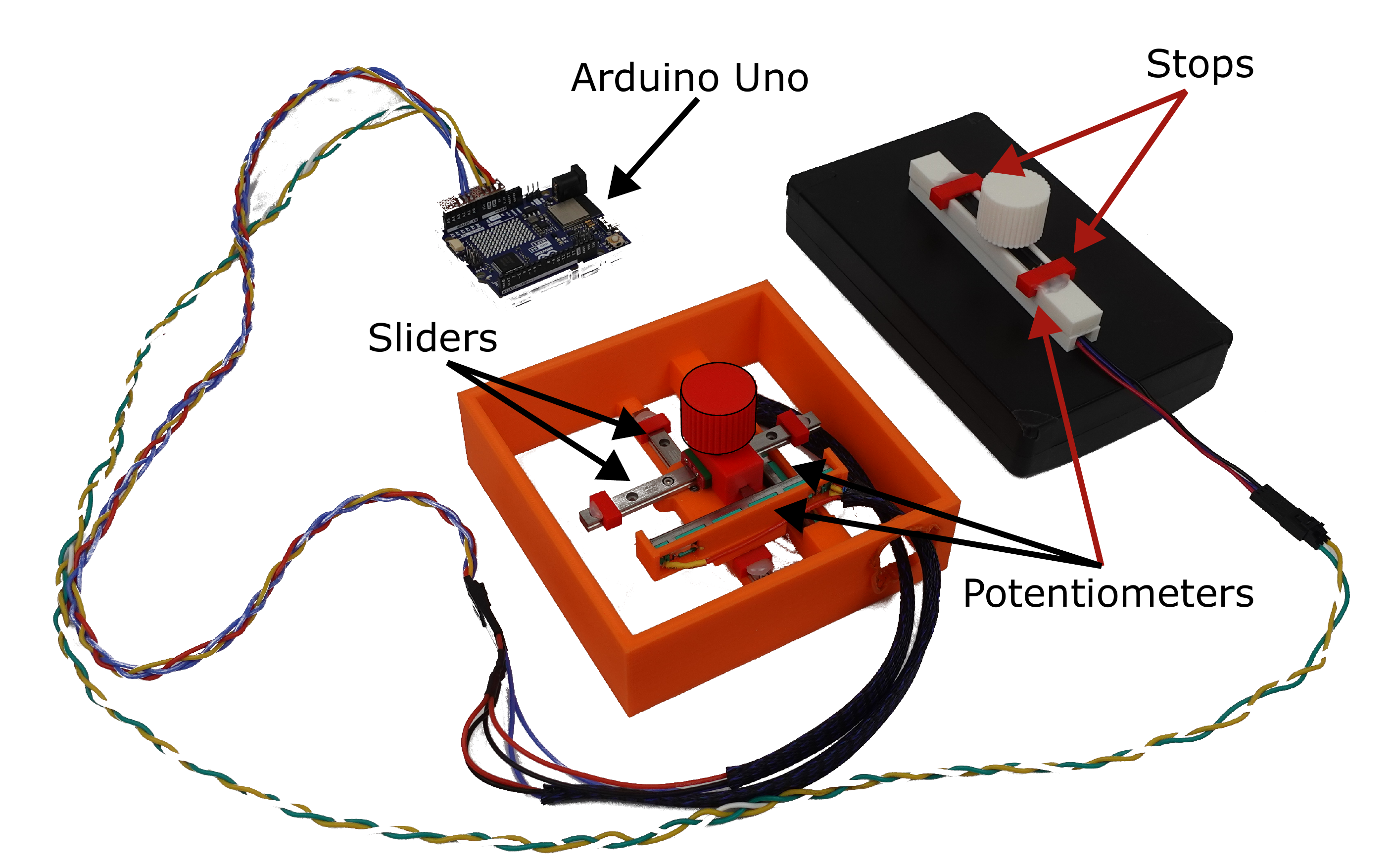}
    \caption{Mechanical sliders for manual parameter tuning. Stops restrict linear potentiometers to their linear region. The sliders consist of rails and low-friction rollers.\label{fig:sliders}}
\end{figure}

As we expect there will be interpersonal variability among user's natural characteristics and preference, we developed a mechanical interface to allow the users to manually adjust the parameters. Tuning interfaces used previously include touch pads~\cite{ingraham2022role} or mechanical dials~\cite{clites2020patient}.
For a neck exoskeleton controlled by eye-movement, tuning interface via a visual display (e.g., touch pads) may not be the best option as a user needs to look at the touch pads to change parameters, which disturbs them from performing visual tasks.
Therefore, we built a system of mechanical sliders that can be felt through the hands without the need to glance over it during parameter tuning (Fig.~\ref{fig:sliders}). 

There are two slider subsystems: a 2-DoF Cartesian system for the deadzone axes and a 1-DoF sliding system for the velocity proportion, for a total of three control parameters. The 1-DoF system is essentially an encased linear potentiometer (B103, TAISS, China). The two-DoF Cartesian system is a handle fixed onto two commercially available prismatic joints with linear bearings, each of which has the same linear potentiometer attached. Each potentiometer has stops to confine the sliders inside the linear region of the potentiometer, giving them a range of motion of about 3 inches per slider. The frame was made from 3D-printed PLA parts. The analog-to-digital converter (ADC) on the microcontroller (Arduino UNO R4 WiFi) was used to convert potentiometer signals to digital values, which was then sent to the computer via USB at a \change{115,200} baudrate. The ADC on the microcontroller has a resolution of 10 bits, but the potentiometer with the smallest linear range brings the resolution down to 982 values.

\section{Experiments with Human Users}\label{sec:exp}

To test our deadzone hypothesis and justify the \change{vector} gaze control model, as well as evaluate the performance of the controller in a human-robot context, we performed a human experiment.
Eight healthy participants without known history of head-neck injuries were enrolled in this study (4 male, \change{4 female}, 23-29 yrs). Four participants wear glasses and were given the option to not wear them during the experiment. Three opted for that option and were able to perform the given tasks. This experiment was approved by the Institutional Review Board at the University of Utah.
Informed consent was obtained from each participant prior to the experiment.

\begin{figure}[htb]
    \centering
    \includegraphics*[width=\linewidth]{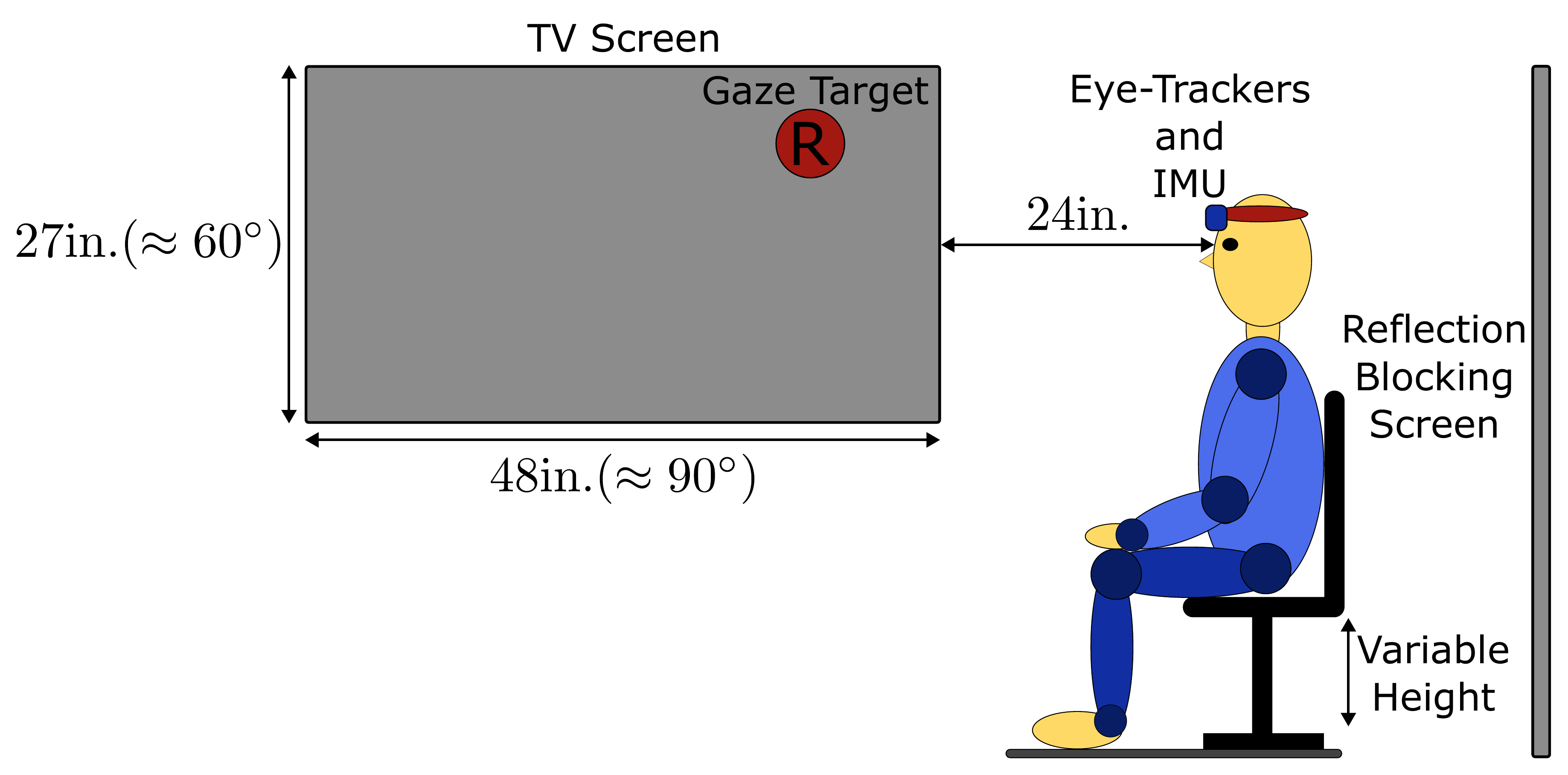}
    \caption{Setup for the eye-head coordination model justification experiment. The IMU and eye-trackers were attached to the participant with a form fitting headband. The reflection blocking screen wrapped around the experiment space to block distracting motions and reflections.\label{fig:exp_setup}}
\end{figure}

This is a three-part experiment. In part 1,
participants were seated in a fixed chair and snugly strapped to discourage torso movements throughout the experiment. A TV screen was placed two feet in front of the participant to fill the field of vision and the seat was adjusted so that when the participant looked straight ahead, their gaze points fell at the center of the TV screen. A screen was placed behind and around the participant to block potential distracting reflections on the TV screen (Fig.~\ref{fig:exp_setup}).

Participants were asked to perform a series of gaze shifts naturally to follow targets presented on the TV screen. The gaze targets were randomly sampled without replacement from an evenly-spaced grid of 200 targets (20$\times$10) spanning 90$^\circ$ horizontally and 60$^\circ$ vertically in the field of view. For each gaze target, the participant was asked to first align the gaze on the center of the screen, with both head and eyes pointing forward. A randomly generated gaze target, along with a randomly generated centered alphanumeric character within the target, was then presented for 2 seconds. The participant was instructed to shift the gaze to look at it in a comfortable manner and name the character out loud. The participant then returned the gaze and head back to the center of the screen for the next target. The purpose of the character was to introduce a cognitive factor to the task for better engagement~\cite{nakashima2014we}.

Head orientations were measured using an inertial measurement unit (IMU) (BNO055, Adafruit) sampled at 40~Hz using a microcontroller (Arduino Uno). The IMU was mounted on the exoskeleton headband (removed from the rest of the exoskeleton) that was placed at the same location on the head as the neck exoskeleton. \change{In all portions of the experiment, the IMU was only used for to measure head orientations as an independent ground truth measurement and not for controlling the neck exoskeleton.} The data were segmented into 200 gaze shift episodes associated with their targets.

For each participant, the deadzone was then calculated as the best-fit central ellipse that classified the most targets correctly. This was performed by grid-searching over the lengths of the vertical and horizontal axes of the deadzone ellipse. 140 values were searched over for the horizontal length, and 100 values were searched over for the vertical length, giving a resolution of about $0.6^\circ$ in each direction.
The velocity proportion was also calculated for each participant after removing the non head moving targets. This was done by calculating the slope of a regression line between the peak head velocity and the angular displacement of the head for each gaze shift. 


In the second part of the experiment, participants repeated the same 200 gaze shifts task, with reshuffled positions and characters, while wearing the exoskeleton controlled by their gaze. \change{The controller's deadzone and velocity proportion were set to the deadzone and velocity proportion values computed from the first part of the experiment}. The main purpose of this part was to evaluate the robotic system's ability to execute the \change{vector} gaze control strategy.
The exoskeleton was attached to each participant at the shoulders and forehead. When a target was presented for 2 seconds, the exoskeleton moved the head to that target according to the control law. Between targets, a homing process was performed, which was achieved by a position controller to realign the head to face the center of the TV screen.

We quantified the accuracy of the robot in three metrics. The first metric is to count the number of targets for which the robot did or did not actuate when its controller dictated otherwise. To do this, we projected the controller's deadzone, that calculated from part 1 of the experiment, and the gaze targets onto the view of the world camera of the eye-tracker. The gaze targets were classified into head moving versus non head moving, and the number of incorrectly located gaze targets relative to the deadzone were counted.
The second metric is to compute the velocity proportion of the robot through a second linear regression and compare it to that measured in part 1 of the experiment. The third metric is to compare the measured and desired orientation of the robot at the end of each gaze shift. Non head moving and inhibited targets were excluded for these comparisons. The desired head orientations were estimated by first taking the largest eye-angle deviation as the position of the gaze target. The controller sought to close the gap between the gaze target and the closest point of the deadzone border. The result of that desired rotation is the desired head orientation. 

In the last part of the experiment, participants tuned the controller parameters with the sliders while performing a different gaze shifting task. This part was designed to demonstrate the sliders use for personalizing the gaze controllers. 
In this part, still wearing the exoskeleton, participants alternated between tuning the controller with the sliders, taking their hands off the sliders, and performing five gaze shifts along a radial direction. The gaze shifts were performed in the same way as the other parts of the experiment. The directions were one of the eight cardinal directions, and each direction was tested twice consecutively, resulting in a total of 80 targets to tune the controller parameters for each participant. The five gaze targets along each direction were distributed in increments of 5$^\circ$ from the gaze center with a Gaussian noise added, which has a standard deviation of 2$^\circ$, to provide more variety of experience in the testing process.

The 1-DoF slider was used for tuning the velocity proportion, while the 2-DoF sliders were used for tuning the horizontal and vertical deadzone diameters. On one end of the sliders, the velocity proportion was zero and the deadzone was at its largest (i.e., the exoskeleton does not move the head for any gaze shifts). The sliders were initially positioned at these ends to ensure safety. On the other ends, the velocity proportion was at its maximum, and the deadzone was at its smallest (i.e., the exoskeleton moves quickly for small gaze shifts). Participants were made aware of these conditions; however, they were blinded as to the ranges and the specific values the sliders commanded. The ranges for horizontal deadzone diameter, vertical deadzone diameter, and velocity proportion were 1$^\circ$ -- 65$^\circ$, 1$^\circ$ -- 45$^\circ$, and 0 -- 2.5 radians per second per radian, respectively.

Slider positions were recorded throughout this part of the experiment, and their final values were compared with the parameter values calculated from the first part. A short qualitative survey was also conducted to gain understanding of the participant's opinion regarding the form factor and experience with tuning the sliders. \change{A short list of questions in Table~\ref{tab:survey} were used to prompt the user's response to various aspects of the slider experiment.} Slider performance from the third part of the experiment was analyzed qualitatively through the survey responses. \change{We also collected other open comments from the participants regarding device comfort, safety, and naturalness throughout the experiment.}

\begin{table}[htb]
    \caption{survey questions after the third part of the experiment \label{tab:survey}}
        \renewcommand\arraystretch{1.5}
        \begin{tabulary}{\linewidth}{L}
        \toprule
        Were the sliders smooth and easy to manipulate?\\
        Do you feel like you could easily explore the various possibilities of the parameters?\\
        Was having three parameters to change too confusing?\\
        Do you feel like you were able to find an acceptable tuning of parameters?\\
        \bottomrule
        \end{tabulary}
\end{table}

\section{Results}\label{sec:result}
\subsection{Model Justification Results}
\begin{figure*}[htb]
    \centering
    \includegraphics*[width=0.75\linewidth]{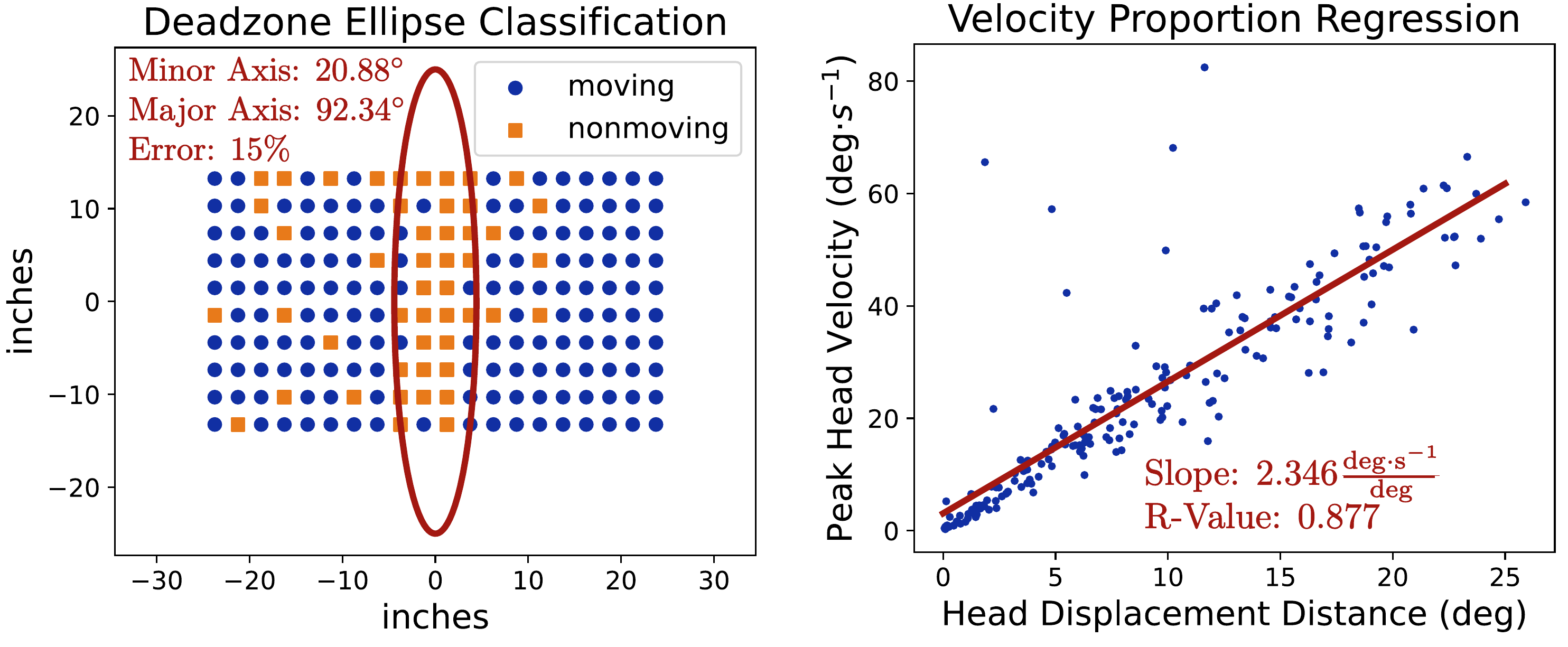}
    \caption{Natural characteristics for participant 4. (left) Deadzone best-fit ellipse. The squares and dots respectively represent the non head moving and head moving targets evenly distributed across the screen. Target locations relative to the center of the screen are measured in inches. The red ellipse is the best-fit ellipse projected onto the screen. Its axis lengths are reported along with its gaze target classification error. (right) Velocity proportion regression. Each sample represents a single head moving gaze shift. The red line is the regression result whose slope is the velocity proportion. \label{fig:natural}}
\end{figure*}

The natural velocity proportion and deadzone characteristics are summarized by participant in Table~\ref{tab:natural_results}. Reported is the slope and \mbox{r-value} of the regression results for calculating the velocity proportion ($p<0.001$ for each participant's regression) and the horizontal and vertical deadzone diameter along with its classification error. The results presented in the velocity proportion columns reaffirm results found in the literature~\cite{freedman1997eye, freedman2008coordination, barnes1979vestibulo}, while the deadzone columns provide new results in eye-head coordination. A wide variety of best-fit ellipse dimensions were observed. As an example, the best-fit ellipse and velocity proportion regression results are shown for participant 4 in Fig.~\ref{fig:natural}.

\begin{table*}[t]
\caption{natural eye-head characteristics measured in the deadzone model justification experiment \label{tab:natural_results}}
\begin{center}
\begin{tabularx}{\linewidth}{cD{.}{.}{1.3}D{.}{.}{1.3}*{3}{D{.}{.}{2.2}}}
\toprule
& \multicolumn{2}{c}{Velocity Proportion Regression} & \multicolumn{3}{c}{Deadzone Best Fit Ellipse}\\
\cmidrule(lr){2-3}\cmidrule(l){4-6}
 Participant & \multicolumn{1}{X}{\centering Slope ($\frac{\text{rad}/\text{s}}{\text{rad}}$)} & \multicolumn{1}{X}{\centering R-Value} & \multicolumn{1}{c}{Horizontal Diameter ($^\circ$)} & \multicolumn{1}{c}{Vertical Diameter ($^\circ$)} & \multicolumn{1}{c}{Classification Error (\%)}\\
 \midrule
1 & 2.298 & 0.946 & 53.16 & 55.53 & 26.50\\
2 & 2.752 & 0.928 & 94.47 & 75.12 & 9.50\\
3 & 1.941 & 0.926 & 7.10 & 77.27 & 16.00\\
4 & 2.346 & 0.877 & 20.88 & 92.34 & 15.00\\
5 & 2.221 & 0.767 & 19.76 & 49.00 & 13.00\\
6 & 2.431 & 0.913 & 83.83 & 66.68 & 19.00\\
7 & 2.553 & 0.966 & 95.01 & 72.15 & 19.50\\
8 & 2.506 & 0.853 & 48.39 & 35.96 & 16.00\\
\bottomrule
\end{tabularx}
\end{center}
\end{table*}

\begin{figure}[htb]
    \centering
    \includegraphics*[width=\linewidth]{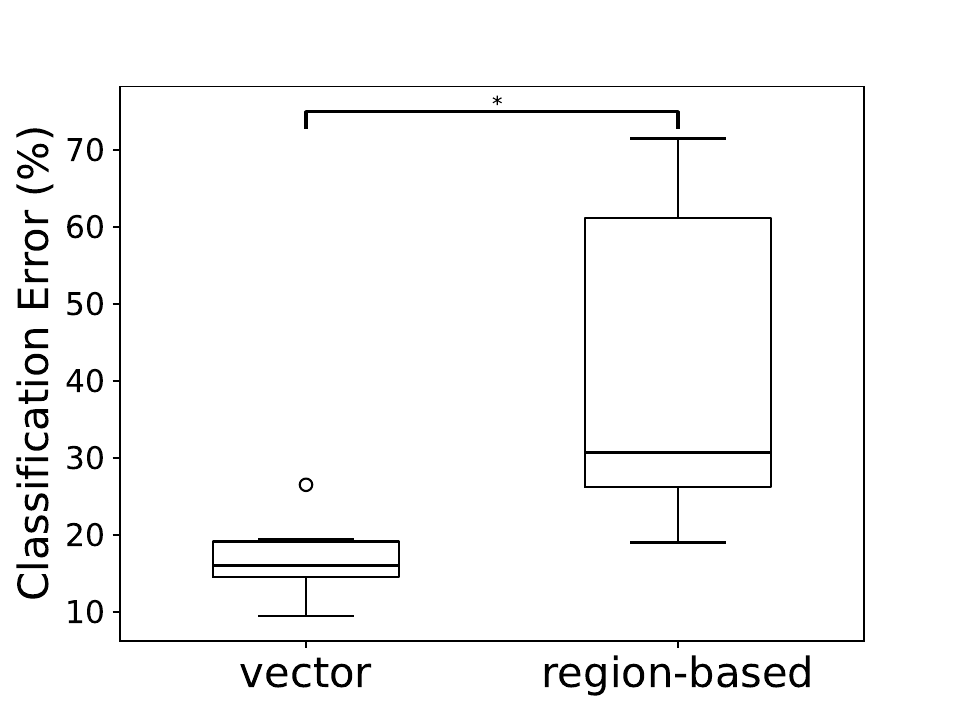}
    \caption{Difference of classification errors for \change{the vector} and region-based deadzones with mean errors of \change{16.81 and 40.44} percent respectively. A Welch's t-test gives a p-value of \change{0.014} indicating the difference is significant. \label{fig:deadzone_diff}}
\end{figure}

To further illustrate the differences between the \change{vector} and region-based controller's deadzones, Fig.~\ref{fig:deadzone_diff} shows box plots of the deadzone classification errors for each controller respectively. The \change{vector controller} deadzone's classification values are those in the last column of Table~\ref{tab:natural_results} and the region-based deadzone classification errors are calculated \change{in the same way but} with a deadzone similar to the previous work~\cite{chang2020bio}, \change{an ellipse of $6.48^\circ$ vertical radius and $12.36^\circ$ horizontal radius} (these dimensions were obtained from the author directly). The mean errors for the \change{vector} and region-based deadzones are 16.81 and \change{40.44} percent, respectively. The classification difference between the two models is statistically significant \change{(p$=$0.014)}, according to a Welch's t-test with the statistical significance level set at 0.05.

\subsection{Robot System Validation Results}
\begin{figure*}[htb]
    \centering
    \includegraphics*[width=0.75\linewidth]{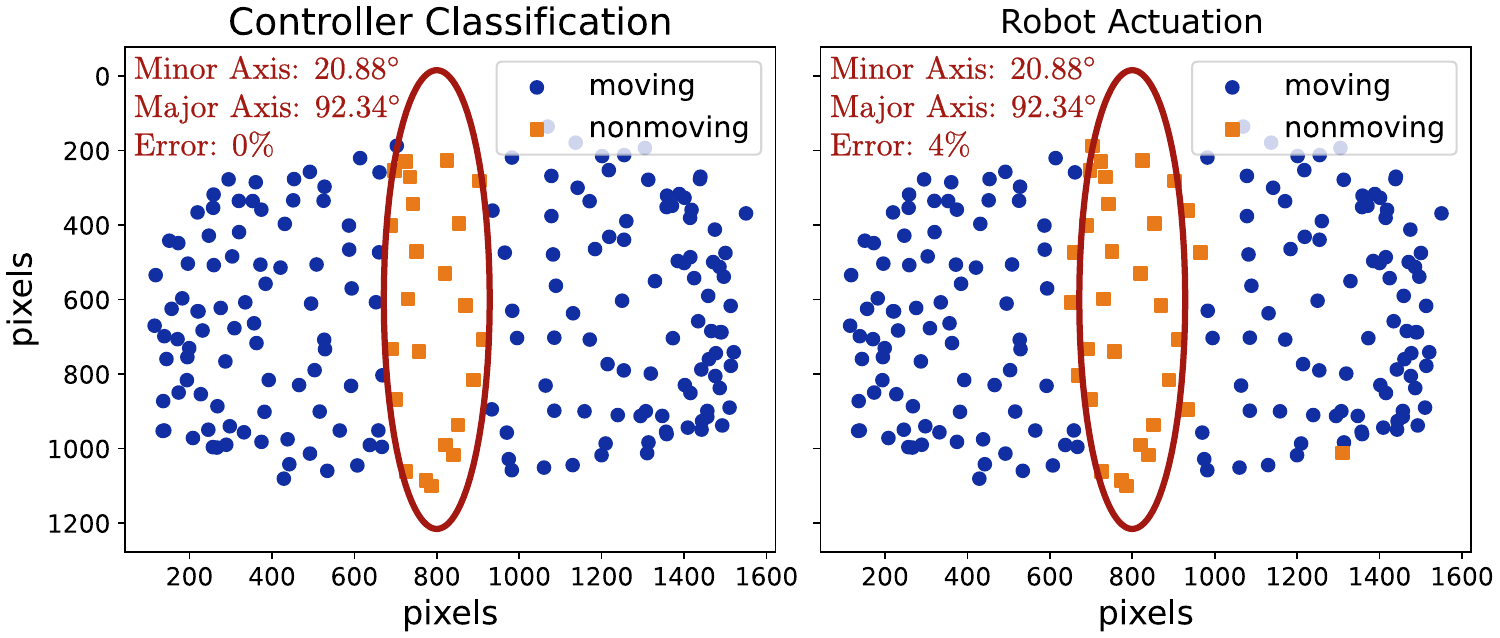}
    \caption{Gaze target classification and robot actuation in the world view camera for participant 4 during the second part of the experiment. Gaze targets are no longer in a grid due to errors such as camera distortion. Target locations in the world camera are measured in pixels. The red ellipse is the natural deadzone projected onto the world camera view. Its axis lengths are reported along with its respective error. (left) Controller classification of gaze targets based on relative position of the deadzone in software. Note that classification is perfect. (right) Actuation performance of the robot in correctly moving or not moving according to controller decisions. Note the robot incorrectly didn't actuated for some gaze targets. \label{fig:pre_post_imu}}
\end{figure*}

Table~\ref{tab:com_robot} summarizes the actuation errors of the exoskeleton, the slope and \mbox{r-value} of the exoskeleton's velocity proportion regression analysis, and the absolute mean and standard deviation differences between the measured and desired \change{lateral bending}, \change{flexion}, and \change{rotation} values of the final orientation. Robot actuation errors from participant 4 are shown as an example in Fig.~\ref{fig:pre_post_imu}. The average time delay between eye saccade and exoskeleton induced head motions was approximately 0.30 seconds.

\begin{table*}[t]
    \caption{performance measures of the exoskeleton from the second part of the experiment \label{tab:com_robot}}
    \begin{center}
        \begin{tabularx}{\linewidth}{cD..{2.1}*{2}{D..{1.3}}*{3}{D..{2.3}@{ }l}}
            \toprule
             & \multicolumn{1}{c}{Deadzone Ellipse} & \multicolumn{2}{c}{Velocity Proportion Regression} & \multicolumn{6}{c}{Absolute Mean(std) Difference End Orientation}\\
             \cmidrule(lr){2-2}\cmidrule(lr){3-4}\cmidrule(l){5-10}\\
            Participant & \multicolumn{1}{c}{Performance Error (\%)} & \multicolumn{1}{X}{\centering Slope ($\frac{\text{rad}/\text{s}}{\text{rad}}$)} & \multicolumn{1}{X}{\centering R-Value} & \multicolumn{2}{X}{\centering Bending ($^\circ$)} & \multicolumn{2}{X}{\centering Flexion ($^\circ$)} & \multicolumn{2}{X}{\centering Rotation ($^\circ$)}\\
            \midrule
            1 &  4.0 & 1.801 & 0.958 & 3.963 & (3.757) & 2.484 & (2.764) & 7.451 & (6.926) \\
            2 & 14.5 & 1.980 & 0.971 & 4.869 & (4.317) & 1.626 & (1.534) & 4.762 & (4.831) \\
            3 &  3.5 & 2.045 & 0.926 & 6.501 & (4.065) & 5.408 & (4.655) & 11.183 & (13.426) \\
            4 &  4.0 & 2.355 & 0.965 & 6.726 & (4.482) & 3.987 & (3.937) & 8.750 & (7.627) \\
            5 &  5.5 & 2.005 & 0.924 & 5.544 & (3.920) & 6.258 & (5.910) & 9.098 & (10.685) \\
            6 & 10.5 & 2.000 & 0.988 & 2.508 & (2.098) & 1.211 & (2.568) & 1.447 & (1.887) \\
            7 & 11.5 & 1.801 & 0.958 & 1.201 & (1.009) & 1.438 & (1.272) & 2.122 & (2.540) \\
            8 & 18.5 & 1.973 & 0.931 & 5.183 & (3.791) & 13.226 & (9.226) & 12.395 & (9.536) \\
            \bottomrule
        \end{tabularx}
    \end{center}
\end{table*}

\begin{figure}[htb]
    \centering
    \includegraphics*[width=\linewidth]{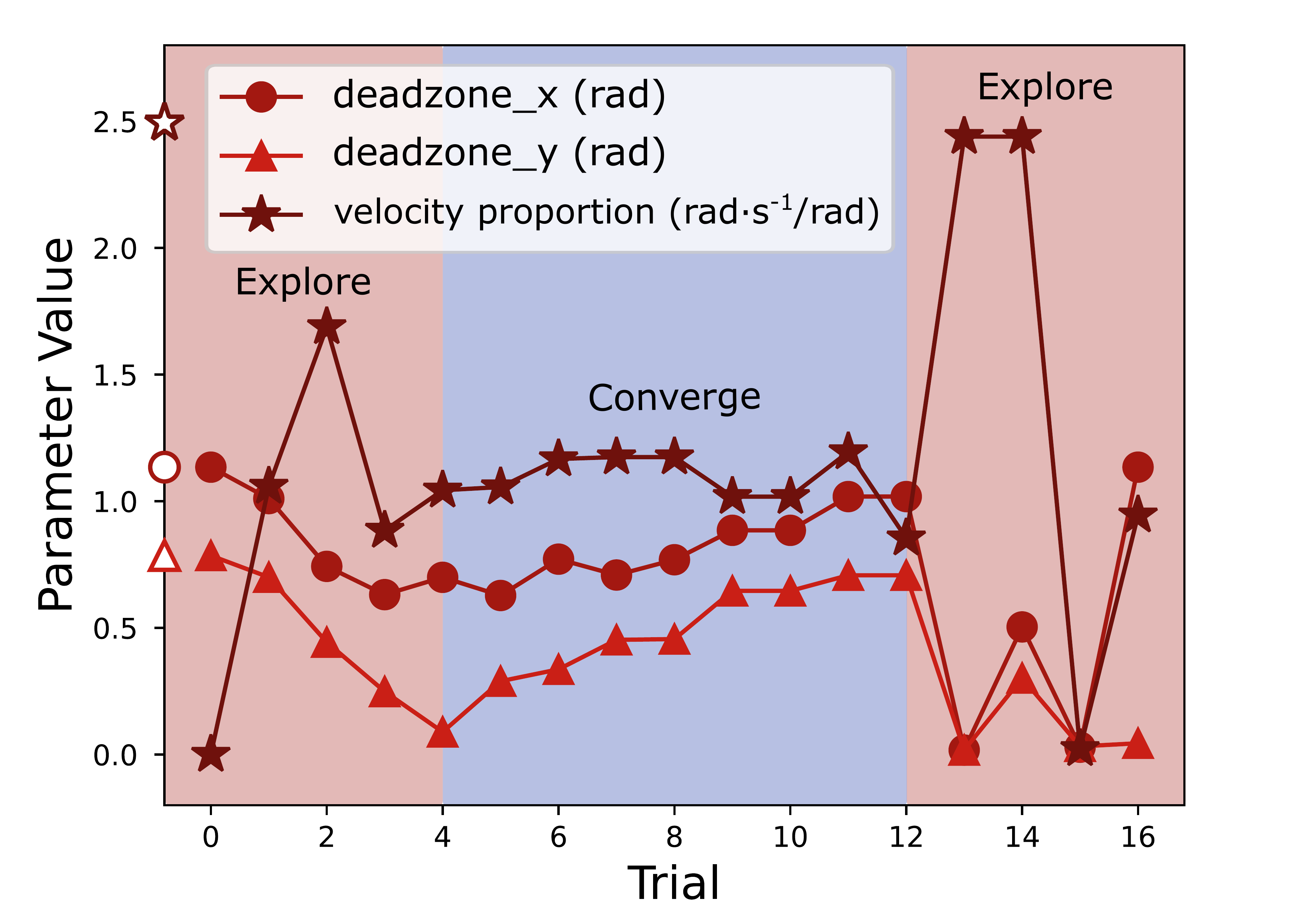}
    \caption{Convergent tuning of parameters for participant 2. The participant explored till the 4th trial, then converged the parameters until the 12th trial. After convergence, the participant explored further. The decorators on the y-axis indicate the maximum tunable values. The deadzone parameters are tuned by the two degree-of-freedom Cartesian system and the velocity proportion parameter is tuned by the one degree-of-freedom system. There are 16 trials.\label{fig:light_train}}
\end{figure}

Responses to the slider questionnaire (Table~\ref{tab:survey}) from all participants were favorable and agreed upon between participants, except for participant 7 who only felt somewhat capable of finding an acceptable tuning of parameters. No difficulties in manipulating the sliders were observed or brought up by participants. Some participants expressed a learning curve with the sliders. General signs of convergence of parameters was observed in all participants; some performed additional tuning exploration after the initial parameter convergence. As an example, the slider parameters of participant 2 during the experiment are shown in Fig.~\ref{fig:light_train}. This participant at first explored, then converged to a set of parameters. Interestingly, this participant then performed another exploration of parameters to their extreme ranges after the convergence.

\begin{figure}[htb]
    \centering
    \includegraphics*[width=\linewidth]{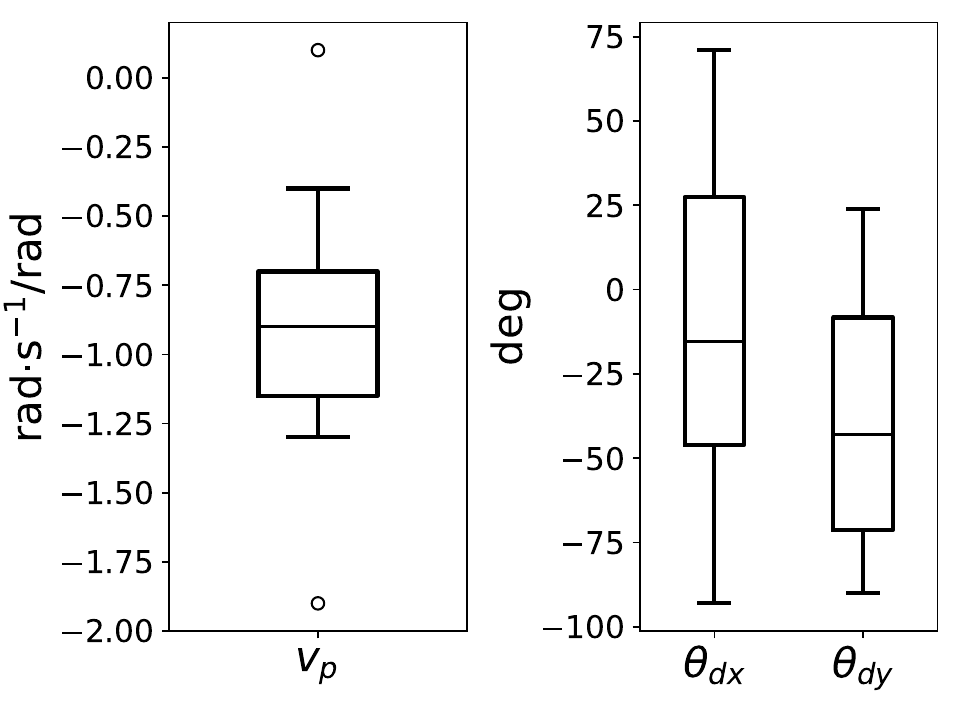}
    \caption{Controller parameter values obtained from the sliders subtracted by the values obtained by the measurements from the first part of the experiment (Table~\ref{tab:natural_results}). The difference of velocity proportion are shown on the left ($v_p$). The deadzone horizontal and vertical diameters are shown on the right \change{($\theta_{dx}, \theta_{dy}$)}. The mean absolute values from left to right are 0.9 radians per second per radian, 44$^\circ$, and 46$^\circ$. \label{fig:slider_diffs}}
\end{figure}

We compared the converged parameter values from the sliders with those fitted using the measured data from part 1 of the experiment. Convergence was determined qualitatively by participant's verbal responses (e.g., ``I am currently satisfied with the parameters'') and by judgments when only small changes of parameters between targets were made. As shown in Fig.~\ref{fig:slider_diffs}, slider tuned values are mostly lower than their measured counterparts with mean absolute values of 0.9 radians per second per radian, 44$^\circ$, and 46$^\circ$ for the velocity proportion and horizontal and vertical deadzone diameters, respectively. Thus, slider tuned values are in disagreement with the natural values that were measured.

Beyond questions regarding the sliders in Table~\ref{tab:survey}, we also asked the participants about their experience with the neck exoskeleton. Most participants felt that the induced movements from the robot were natural and comfortable; however, some reported increased discomfort over time due to exoskeleton constraints (e.g., pressure from robot fittings, limited range of motion restricted by linkages). Participants were engaged throughout the experiment; no participants reported or were observed to remove their gaze from the targets during the experiment.

\section{DISCUSSION}\label{sec:discuss}
This paper presented a novel control strategy for powered neck exoskeletons using eye movements as input, which was designed according to eye-head coordination behaviors. In this new \change{vector controller} model, angular velocity of the head is proportional to the distance of a visual target from the center of the field of vision. Unlike the previous model (``region-based controller'') which only updates the head orientation in \change{constant strictly} vertical or horizontal \change{step sizes}, the new \change{vector} model preserves the natural velocity behavior of the head to respond to gaze shifts~\cite{freedman1997eye, freedman2008coordination}. Mapping the gaze shifts (which can be thought of as step inputs) with head velocity through a positive gain (velocity proportion) also results in a stable system which ensures safety for head control. The regression results on velocity proportion show strong correlation between the angular displacement of the eyes and the peak angular velocity of the head (Fig.~\ref{fig:natural}). This is in agreement with what has been found in the literature~\cite{freedman2008coordination, barnes1979vestibulo}. 

Additionally, we proposed a novel deadzone model using \change{personalizable} elliptical shapes \change{verified by data}. This deadzone, centered in the field of vision, depicts a region within which gaze shifts trigger no change in head orientation. This region is important for the control of powered neck exoskeletons to classify when to keep the head stationary and when to move the head in response to large gaze shifts. However, this region was not well characterized \change{previously~\cite{chang2020bio}}. \change{Contrary to the belief that the deadzone should shape like a horizontal ellipse given wider horizontal field of view~\cite{strasburger2011peripheral}, our data suggest a wider shape varieties of the deadzones, including horizontal, vertical, and near circular shapes.}

Importantly, this deadzone should not be confused with the effective \mbox{occulo-motor} range (EOMR)~\cite{freedman2008coordination,tweed1995eye}. The EOMR is the region of comfortable eye positions that are used in most situations. Eye angles inside the EOMR may cause head movements, so it is not useful to determine when the head will move. 
Our result shows (Fig.~\ref{fig:deadzone_diff}) that the new deadzone model is more accurate to classify whether a gaze shift triggers the head movement than using the previous \change{elliptical} deadzone with fixed radii for all participants \change{({$p=0.014$})}. 

The shape and size of the deadzone of each participant was found to be unique (Table~\ref{tab:natural_results}). For example, the deadzone of participant 1 is near circular while the deadzones of other participants are horizontal or vertical ellipses. For participants 2 and 7, their deadzone nearly covered the entire screen, indicating their general lack of head movement for the selected head-eye coordination task. Despite efforts to encourage head movements using a task with a cognitive component (i.e., reading out loud the random alphanumerical character), these two participants \change{performed fewer head motions}. In the literature~\cite{fuller1992head, lee2024patterns}, individuals with little movements of the head during visual tasks are referred to as non head movers, and it is possible that these two participants fall in this category. \change{Despite this behavior, the neck exoskeleton was able to mimic their natural behavior by not moving the head for proximal gaze targets, but still moving the head naturally for the more distal gaze targets.}

Each participant was also found to have a unique velocity proportion (Table~\ref{tab:natural_results}). The observed variability of size and shape of the deadzone and velocity proportion supports the need for personalization for these parameters, thereby highlighting a key limitation in the previous ``region-based'' model using fixed parameters.

Our study of the deadzone and velocity proportion is cross-sectional and does not account for how these characteristics might change for different tasks. For example, the EOMR were found to skew for different starting positions for gaze shifts~\cite{smith2014covert}. Perhaps for tasks requiring greater cognition or energy, one might expect a larger velocity proportion and smaller deadzone. The variance of deadzone and velocity proportion is a topic that needs further study to design more natural and task-accommodating controllers.

Due to the low classification errors of the elliptical deadzone and \mbox{r-values} of the velocity proportion, we felt justified in using these key ideas of eye-head coordination in \change{the vector} controller design. However, while controller designs based on natural kinematic behavior are promising, such controllers need to be studied and evaluated in various contexts beyond emulating natural kinematics for laboratory tasks. Such contexts include user's preference or task performance during daily living activities. \change{It is therefore possible that a user may feel a preference towards even the region-based controller instead of the natural vector controller; in other words, a user may view other values to be more important than just the kinematic coupling between the head and eyes. This robotic implementation and methods for personalization provide the necessary tools to study these important research questions.}

The \change{vector} control model was successfully implemented in a neck exoskeleton. As shown in Fig.~\ref{fig:pre_post_imu} (left), no gaze targets were misclassified according to their position in the world camera view. However, the exoskeleton was occasionally unable to execute the controller's planning, as shown in Fig.~\ref{fig:pre_post_imu} (right) where IMU measurement of the head rotation indicated incidences of misclassifications.
Reasons for this error include that gaze targets close to the deadzone border may not cause movement commands big enough for the IMU to recognize or strong enough for the robot to overcome frictions in the hardware to move the head, or that participants may have been fighting against the robot. As the robot is less likely to move for gaze targets close to the border of the deadzone, we observed higher actuation errors for participants who have much larger deadzones (e.g., participants 2 and 7). However, the overall actuation errors were low (\change{Fig.~\ref{fig:pre_post_imu}}), suggesting good tracking ability of the exoskeleton to execute the control model.

We believe that the observed velocity proportion for each participant was an overestimate because about 4\% of the gaze targets were inhibited by the physical frame of the eye-tracker initially and were thus incomparable with the other gaze shifts (i.e., the user must perform searching maneuvers to find these targets). Removing these targets, however, required time-consuming inspection of videos. Therefore, in the interest of participants' time for the later parts of the experiment, these targets were included in its calculation of the parameters during the experiment. 

This was also corroborated by the results from the velocity proportion regression analysis. For all participants, excluding participant 7, \mbox{r-values} were equal to or higher than the natural behavior (Table~\ref{tab:natural_results}).

The just noticeable difference of low head velocities was reported to be about 0.46$^\circ$ per second at lowest and increasing to about 2.26$^\circ$ \change{per second} for moderate head velocities and about 5.16$^\circ$ \change{per second} for higher head velocities~\cite{mallery2010human}. \change{We can calculate the maximum possible difference of head velocity between natural and robotic conditions for each participant by
\begin{equation}
    \Delta_{g}v_{p,n}-\Delta_{g}v_{p,r}
\end{equation}
where $\Delta_{g}$ is the maximum possible deviation of gaze from the respective participant's deadzone, and $v_{p,x}$ is the natural or robot velocity proportions found in Table~\ref{tab:natural_results} and Table~\ref{tab:com_robot}. The largest difference of head velocity was about $22.89^\circ$ per second for participant 8 and the smallest was about $0.46^\circ$ per second for participant 4. Some users will therefore be unlikely to notice any differences of head velocity between their natural state and with the robotic neck brace, while others are more likely to notice such differences.}

With few exceptions we were able to achieve average end orientation accuracies of less than 8$^\circ$, the worst being 13.23$^\circ$ for participant 8 in the \change{flexion} direction. Standard deviations are low for some participants, but higher for others (5$^\circ$ to 20$^\circ$). These differences in precision are likely due to small movements of the attachments on the shoulders and some play observed in the linkages and end effector of the exoskeleton, highlighting the needs for precise hardware fabrication and fitting methods that better conform to body shapes~\cite{brignone2024towards}. Another reason may be that the exoskeleton moves farther and faster for smaller deadzones and thus has more opportunity for errors to accumulate such as a drifting head \change{lateral bending}. \change{Eye tracking only provides two degrees of freedom for control, thus head \change{bending} causes errors in inferring desired head velocities from the gaze.} We did not directly regulate unwanted head \change{bending} in this experiment because the drifting head \change{bending} was minimized with the exoskeleton being reset into the home position for each target. However, this can be regulated in future control implementations by minimizing the measured \change{bending} angle through a closed loop control. 

In the slider tuning portion of the experiment (part 3), participants reported little, if any, physical or cognitive difficulties in using the sliders and felt they were able to find desirable parameters. 
When comparing the slider-tuned parameter values (after a convergence) with those measured from part 1 of the experiment (Fig.~\ref{fig:slider_diffs}), we found that most of the slider values were lower than the measured ones. This means that user-tuned parameters generally have a smaller deadzone but with a lower velocity proportion. Even though the velocity proportion is lower, a smaller deadzone increases the peak head velocity for the same gaze shift\change{, potentially balancing out the change of velocity.}

\change{This difference of parameters could be an indication of change of eye-head coordination behavior due to fatigue or robot interaction, or a reflection of users cautiously tuning for safe parameters. This behavior emphasizes the need for personalization, as the most natural behavior from the robot may not be the most desirable behavior for the user. Indeed, it is possible that the region-based controller may be more preferred than the vector controller, depending on the user's preference. This should be studied in future works to increase understanding of how users interact with these robotic neck braces to provide satisfactory assistance.}

Additionally, as shown in Fig.~\ref{fig:light_train}, we did observe temporary convergence of parameters before new exploration of the parameters in some participants. This may indicate a search strategy that some participants use: as they became more familiar with the tuning and exoskeleton control, they were more likely to explore the entire parameter space.
As we are only concerned with the technical validation of the slider tuning interface in this paper, their role in tunability or personalization of eye-tracking controllers in the context of human preference was not specifically studied. However, our preliminary results motivate research in these directions.
Most importantly, we believe that the slider interface provides an appropriate tool for such future studies.

\change{As we were interested in further refining the slider interface, only general open comments and feedback from the participants were collected. Future studies should provide more quantitative measures of user's perceptive responses through use of Likert-style questionnaire designs.}



Some scenarios required the participants to, in a sense, override the \change{vector} controller to better align the gaze. For example, as we employed a hard border of the deadzone, a gaze target that is inside but close to the border of the deadzone will not trigger a head rotation to center the gaze. To center the gaze, a user would need to look farther away from the target to induce the head rotation towards that target. This limitation could be addressed in the future with a ``soft'' deadzone design that allows small head rotations for scenarios like this. We found that sometimes the gaze saccade went beyond the deadzone but returned quickly back to the center of vision (i.e., a quick glance), which did not trigger motion of the robot. This was not contributed by a filter in the software but mainly due to the robot communication and computational delays (about 0.3 seconds), which is longer than most quick saccade (0.025 -- 0.04 seconds~\cite{freedman2008coordination}). 

\section{CONCLUSION}\label{sec:conclusion}
In this paper, we presented a new gaze control model according to the head-eye coordination during gaze shifts\change{, the vector controller}. We also performed the human-user validations of the robotic implementation of \change{the vector} control model in a powered neck exoskeleton. Furthermore, we evaluated tools for potential personalization of the control model. Our work provided a validated platform to study eye-tracking controllers, human-robot interaction, and personalization and preference of controllers for powered neck exoskeletons. Ultimately, our work advances the translational efforts to restore head-neck motions for patients with neck muscle weakness through powered neck exoskeleton technologies.

\addtolength{\textheight}{-8cm}

\section*{ACKNOWLEDGMENT}
We thank the 8 participants who participated in this study.

\bibliographystyle{ieeetr}
\bibliography{ref}

\end{document}